\documentclass{article}

\usepackage[preprint]{neurips_2022}

\usepackage[utf8]{inputenc} 
\usepackage[T1]{fontenc}    
\usepackage[colorlinks=true,linkcolor=blue,citecolor=blue]{hyperref}%
\usepackage{url}            
\usepackage{booktabs}       
\usepackage{amsfonts}       
\usepackage{nicefrac}       
\usepackage{microtype}      
\usepackage{xcolor}         
\usepackage{graphicx}
\usepackage{natbib}
\usepackage{amsmath}

\usepackage{multirow}

\setcitestyle{authoryear,open={(},close={)}}

\title{Adapting Pretrained Vision-Language Foundational Models to Medical Imaging Domains}

\author{
    Pierre Chambon\thanks{equal contribution} , Christian Bluethgen\footnotemark[1] , Curtis P. Langlotz, Akshay Chaudhari\\
    Center for Artificial Intelligence in Medicine and Imaging \\
    Stanford University\\
    \texttt{\{pchambon,bluethgen,langlotz,akshaysc\}@stanford.edu}
}

\begin{document}

\maketitle

\begin{abstract}
Multi-modal foundation models are typically trained on millions of pairs of natural images and text captions, frequently obtained through web-crawling approaches. Although such models depict excellent generative capabilities, they do not typically generalize well to specific domains such as medical images that have fundamentally shifted distributions compared to natural images. Building generative models for medical images that faithfully depict clinical context may help alleviate the paucity of healthcare datasets. Thus, in this study, we seek to research and expand the representational capabilities of large pretrained foundation models to medical concepts, specifically for leveraging the Stable Diffusion model to generate domain-specific images found in medical imaging. We explore the sub-components of the Stable Diffusion pipeline (the variational autoencoder, the U-Net and the text-encoder) to fine-tune the model to generate medical images. We benchmark the efficacy of these efforts using quantitative image quality metrics and qualitative radiologist-driven evaluations that accurately represent the clinical content of conditional text prompts. Our best-performing model improves upon the stable diffusion baseline and can be conditioned to insert a realistic-looking abnormality on a synthetic radiology image, while maintaining a 95\% accuracy on a classifier trained to detect the abnormality.

\end{abstract}

\begin{figure}[hbt!]
  \centering
    \includegraphics[width=138mm]{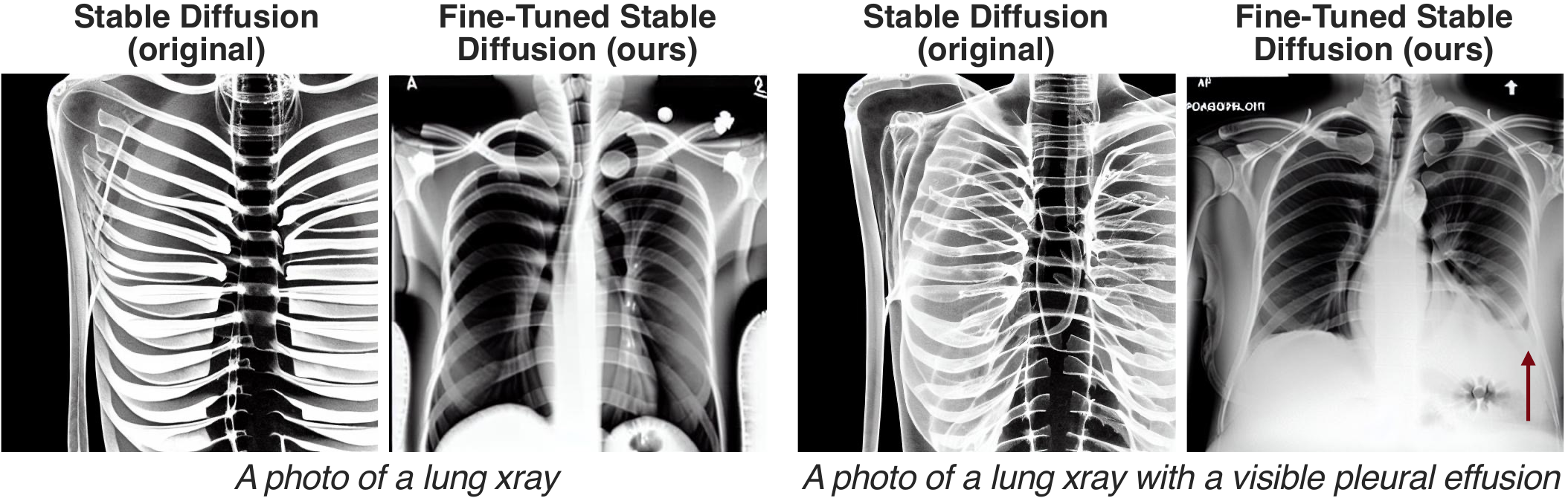}
  \caption{Generated images by both the original Stable Diffusion model and our fine-tuned model on radiology images. The prompts are designed to compare a standard radiology image with no particular findings, and the insertion of the frequently encountered finding "pleural effusion" (red arrow).}
  \label{figure-model-outputs}
\end{figure}

\section{Introduction}

In recent months, latent diffusion models have gained immense popularity by enabling state-of-the-art image generation amenable to fine-grained control of the image generation process at inference time via conditioning of the denoising process (e.g., using text prompts) \citep{ramesh2022, Rombach2022, imagen2022}. Such models, termed foundation models \citep{foundationmodels2021}, have been trained with large, curated multi-modal datasets such as LAION-5B that consists of natural images and their captions \citep{schuhmann2022laionb}. The impressive generative capabilities of such models permit creation of high-fidelity synthetic datasets that may be used to augment traditional supervised machine learning pipelines in scenarios that lack training data. 

One particular area that such an advance would be beneficial in is the domain of medical imaging, where there is a paucity of high-quality labeled datasets. Annotating such medical imaging datasets typically requires trained medical experts who are capable in interpreting subtle, but semantically meaningful, image features. Despite the lack of large curated medical imaging datasets, one benefit that such medical imaging examinations have is that there is typically a text-based radiology report that describes pertinent findings from the imaging study. Leveraging the vision-language understanding capabilities of latent diffusion models could potentially provide an intuitive mechanism to create synthetic medical imaging data by prompting with relevant medical keywords or concepts of interest.

In this study, we explore the representational bounds of large vision-language foundation models and evaluate how to utilize pretrained foundational models to represent medical imaging studies and concepts, despite models never having been explicitly trained on these concepts. We utilize chest X-rays (CXRs) for this study as they are most common imaging modality globally. CXRs are fast to acquire, inexpensive, can provide important patient health insights, and can identify and monitor a variety of pathologies. We explore and quantify the representational capacity of the Stable Diffusion model \citep{Rombach2022} to characterize the efficacy of both its language and vision encoders as applied to CXRs. We further explore different strategies for improving the representational capacity of non-domain-specific foundational models for representing medical concepts specific to CXRs. These experiments help provide novel decision making insights regarding whether such foundational models can accurately represent complex biomedical concepts for clinically-relevant downstream tasks, without explicit training on such concepts. In this study, we specifically show the following:

\begin{enumerate}
    \item The LAION-5B pretrained variational autoencoder (VAE) of the Stable Diffusion pipeline can reconstruct CXR images at arbitrary resolutions out-of-the-box
    \item A frozen CLIP text encoder can generate powerful medical embeddings with sufficient context to allow medically accurate images, in conjunction with the methods below
    \item Replacing the frozen CLIP encoder with a frozen in-domain text encoder with a projection head trained on LAION to map in-domain embeddings to CLIP embeddings, did not result in better images
    \item Textual inversion can be used to learn complex medical concepts like 'pleural effusion' in a few-shot manner
    \item Fine-tuning the U-Net component enables high-fidelity CXR image generation with the capability to insert custom pathologies (see examples in Figure \ref{figure-model-outputs}).
\end{enumerate}

We substantiate our findings using quantitative image quality and classification metrics and verify the results by qualitative and domain-specific evaluation by a thoracic radiologist.

\section{Materials and Methods}
\label{materials-methods}

\subsection{Datasets}
\label{section-datasets}

Two large, publicly available CXR datasets were used to perform the experiments presented in this work. The CheXpert dataset contains 224,316 chest radiographs of 65,240 patients treated at Stanford Hospital between October 2002 and July 2017 in both inpatient and outpatient centers \citep{irvin2019chexpert}. The second dataset, MIMIC-CXR (version 2.0.0), contains a total of 377,110 images from studies performed at the Beth Israel Deaconess Medical Center in Boston, MA, USA under institutional review board approval \citep{Johnson2019}. From each dataset, 1000 frontal (i.e., anterior-posterior or posterior-anterior projection) radiographs were sampled randomly for this study. The images and their associated reports were used for experiments and study of the variational autoencoder and of text encoders. 

Five images with no apparent findings, and five images featuring clearly visible pleural effusion as sole finding were manually selected. Unusually cropped or colorized images were discarded. The selected images were paired with a set of simple, synthetically generated prompts to form image-text pairs used for fine-tuning the Stable Diffusion components with various approaches.

Finally, a sample of one million text prompts from the LAION-400M dataset 
was used for textual projection training and experiments.

\subsection{Stable Diffusion}

\begin{figure}
  \centering
    \includegraphics[width=130mm]{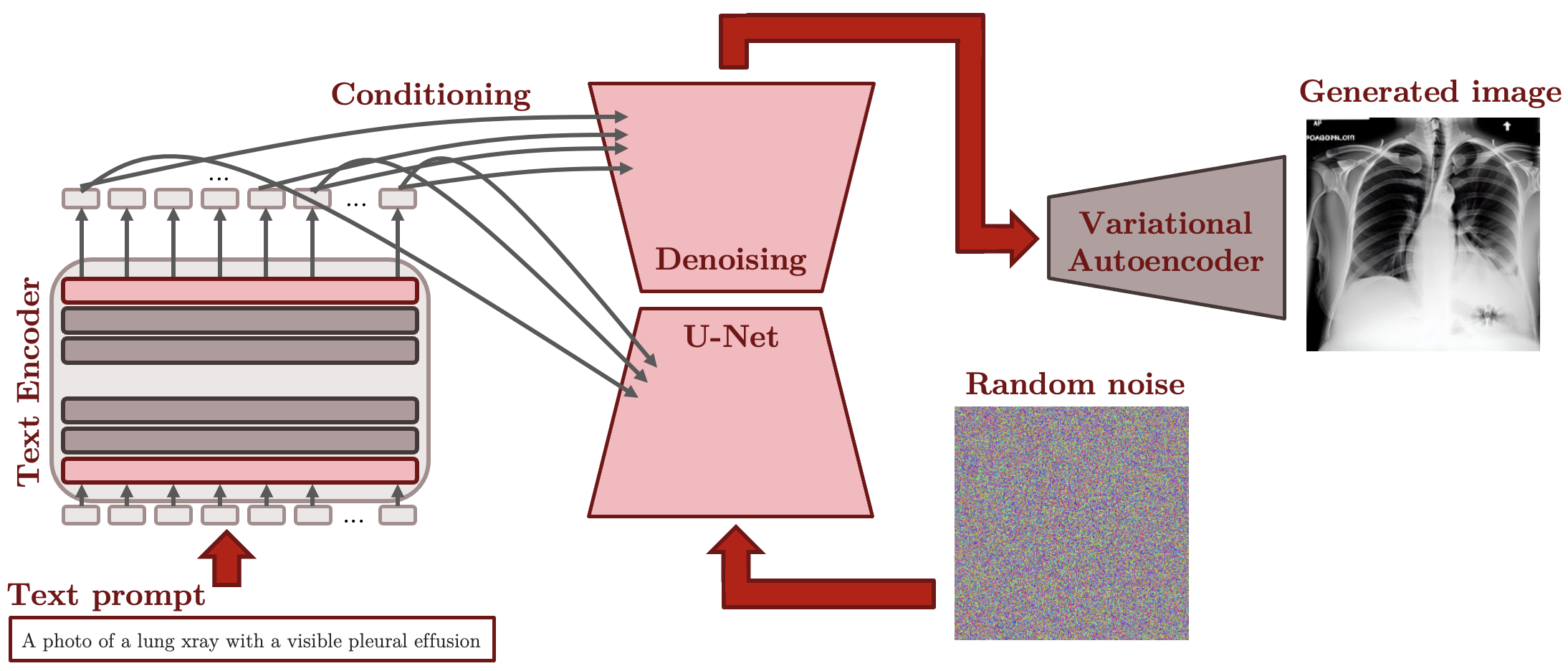}
  \caption{Stable diffusion architecture, run in the radiology setting to generate synthetic radiology images.}
  \label{figure-stable-diffusion}
\end{figure}

The Stable Diffusion pipeline (depicted in Figure \ref{figure-stable-diffusion}) is accompanied by a CLIP text encoder that parses text prompts to create a 768-dimensional latent representation \citep{clip2021}. This latent text representation is used to condition a denoising U-Net to generate images in the latent image space using random noise as initialization. Finally, the decoder component of a VAE is used to map this latent representation to the pixel space. While the original generative model has been trained with image and text captions arising from natural imaging domains, the extent of its capabilities for representing medical concepts and images remains unclear.  
To adapt the stable diffusion model for in-domain image generation, especially for radiology images and prompts, we can leverage each component and train it, or not, depending on its capabilities to represent in-domain data. More particularly, we can assess:

\begin{itemize}
    \item Whether the variational autoencoder (VAE) alone is capable of reconstructing radiology images without losing general visual aspect or, more importantly, clinically significant features.
    \item Whether the text encoder alone is capable of projecting clinical prompts to the text latent space while preserving clinically important information.
\end{itemize}

Section \ref{variational-autoencoder} presents the methods used to assess the reconstruction quality of the VAE, assessing whether it requires in-domain fine-tuning; Section \ref{method-text-encoder} describes the experiments researching the quality of the CLIP text encoder and other in-domain text encoders; and Sections \ref{textual-projection}, \ref{textual-projection}, \ref{u-net-fine-tuning} present methods to fine-tune various components of the stable diffusion model for the radiology domain.

\subsection{Variational Autoencoder}
\label{variational-autoencoder}

As latent diffusion model, Stable Diffusion translates image inputs into a latent space before performing the generative denoising process, using an encoder trained to remove high-frequency details representing perceptually negligible features (“perceptual compression”)\citep{Rombach2022}, before learning to model the semantic and conceptual composition of the data necessary for the actual generative process. To analyze how well medical imaging information is preserved while passing through the VAE, CXR images sampled from CheXpert or MIMIC (“originals”) were encoded to latent representations and reconstructed into images (“reconstructions”). To this end, the pretrained VAE was used without additional fine-tuning, sampling from the generated latent distribution as provided by the Stable Diffusion pipeline VAE encoder, and decoding the latent representations without additional processing, using the VAE's decoder component.

Reconstruction quality was quantitatively assessed by calculating the root-mean-square error (RMSE), the peak signal-to-noise ratio (PSNR) and the structural similarity index measure (SSIM) for each image-reconstruction pair. Additionally, the Fréchet inception distance (FID, underlying model: Inception V3, 2048 features) was calculated on batches (batch size = 32) to compare the distribution of reconstructions to the distribution of original images\citep{szegedy2015, heusel2017}.

Qualitatively, the reconstruction quality compared to the original image input was assessed by a radiologist with 7 years of experience in reading CXR studies, using a scoring system ranging from 1 to 5 (5: Very good reconstruction with essentially non-inferior diagnostic quality to the original, 4: Good reconstruction with noticeable errors not negatively influencing diagnostic quality, 3: Moderate reconstruction errors with possible negative effects to diagnostic performance, 2: Severe reconstruction errors or errors of any level leading to hallucinated lesions, 1: Severe reconstruction errors yielding the image nondiagnostic) on 100 randomly sampled original-reconstruction pairs. 

The effect of the reconstruction process on classification performance was analyzed using a model pretrained to detect 18 different pathologies commonly encountered in CXR (DenseNet-121, torchxrayvision library, version 0.0.37)\citep{Cohen2022xrv}\citep{Cohen2020multids}. Classification accuracy, area under the receiver operator characteristic curve (AUC), and F1 score were calculated for 12 of the labels included in CheXpert, MIMIC-CXR and the pretrained model. For this step, uncertain findings (='-1') were considered positive findings (='1'), while missing values were treated as absence of the corresponding finding (='0'). Additionally, latent representations of original and reconstructed images derived from intermediate layers of the classification modelwere compared by calculating their pairwise cosine similarity.

\subsection{Text Encoder}
\label{method-text-encoder}

In the domain-specific setting of radiology reports and images, the goal of this project is to be able to condition the generation of images on associated medical conditions, that can be represented through a text prompt (e.g., in the form of a report). Therefore, the capability of the text encoder to correctly represent medical features in the latent space is critical for the rest of the Stable Diffusion process, in particular the U-Net operating in the latent space, to be able to generate images that are anatomically correct and representing the correct set of abnormalities. 

A set of potential text encoders that could be interesting to accurately represent medical features was found through literature research and study of previously published pre-trained language models in the field: PubMedBERT \citep{Gu_2022}, ClinicalBERT \citep{https://doi.org/10.48550/arxiv.1904.05342}, SapBERT-from-PubMedBERT-full text \citep{liu-etal-2021-self}, RadBERT (huggingface.co/StanfordAIMI/RadBERT), CXR-BERT-general \citep{https://doi.org/10.48550/arxiv.2204.09817}, CXR-BERT-specialized \citep{https://doi.org/10.48550/arxiv.2204.09817} and finally the Clip text encoder \citep{https://doi.org/10.48550/arxiv.2103.00020}.

Leveraging publicly available datasets of CXRs with accompanying free-form text reports (see Section \ref{section-datasets}), radiological image-text pairs can be defined. Additionally, corresponding image-level abnormality labels can be extracted from the reports using the CheXpert labeler model \citep{irvin2019chexpert}. Then, for each particular $text\_encoder$ model and the corresponding $report\_list$ of elements $report$, one can run the $report$ through the model and get a representation $text\_encoder(report)$. Nevertheless, there exist several ways to extract embeddings from these text encoders, all based on a transformer architecture: extracting the last layer hidden state of the associated CLS token, "CLS hidden state"; extracting the last layer hidden states of each tokens and averaging these representations, "mean hidden states"; using the pooler output, "pooler output"; Using a model specific extraction method, if available, "model specific".

The combination of a $text\_encoder$ model and the associated extraction method $extraction$ gives a function $extraction \circ text\_encoder$ that takes an input report and outputs a document-level representation. This way, for a defined $text\_encoder$ model and $extraction$ method, one can obtain document-level embeddings of radiology reports and assess the quality of these embeddings and thereby the capability of a text-encoder to encode medical content.

For the evaluation, we first obtain the document-level embeddings on the impression section of each radiology report, obtained through regex parsing. This gives:
$$
impression\_embeddings = extraction \circ text\_encoder(impression_sections)
$$

For all the text-encoders that we study, the latent representations are of dimension 768. Therefore for 700 impression sections, $impression\_embeddings$ is a $700 \times 768$ matrix. 

Then, we can compute the impression-impression similarities in the latent space $$similarities = impression\_embeddings \times  impression\_embeddings^{T}$$
We then compute a metric, that we denote the $CheXpert@k$ metric, that for each report $i$ find the $k$ most similar reports, and then measure the proportion of reports that share the same CheXpert label. If $chexpert\_labels$ is a list of the chexpert labels corresponding to the reports, we have:
$$
CheXpert@k_i = \frac{sum(chexpert\_labels[argsort(similarities[i])[-k:]]==chexpert\_labels[i])}{k}
$$

And then over all reports we get: 
$$
CheXpert@k = \frac{\sum_{i=1}^n CheXpert@k_i}{n}
$$

Notice that in the implementation of this metric, a filter is added to $CheXpert@k_i$, so that among the k most similar reports, the report being compared to is not retrieved. In addition, the metric $CheXpert@k$ can be computed over each class instead: so for each abnormality class, we can average the $CheXpert@k_i$ scores, where the similarities are still computed over the reports of all classes. A macro-averaged score can then be retained for comparison purposes.

\subsection{Textual Projection}
\label{textual-projection}

Building upon the Stable Diffusion work, as a first experiment to generate domain-specific images, we replace the CLIP text encoder, which was kept frozen during the original Stable Diffusion training, with a domain-specific text encoder pretrained on biomedical or radiology data. The rationale for this architectural change is to leverage the presumed higher domain-specific representational capacity of the alternative text encoder to provide more useful latent representations of the medical information stored in radiology reports for the downstream generative process.

Simply replacing the CLIP text encoder with a different encoder would arguably lead to catastrophic performance, given that latent spaces can be structured in a very different manner. There is no guarantee that any latent feature is redundant between two different text encoders. We therefore propose to train a projection capable of translating, in part, the latent representations of one text encoder to another, with the idea that running a radiology prompt through the in-domain text encoder, and then translating its generated latent representations through this trained projection, could allow embeddings to be sufficiently well aligned to be able to be used as conditioning tool for the U-Net while providing enhanced clinical representations.

To train this projection, we use the LAION-400M dataset and define a $projection$ as an MLP model, taking a 768-dimensional input and mapping it to a 768-dimensional translated output. As a first approach, we take $projection = Linear \circ ReLU \circ LayerNorm \circ Linear$ and train it using MLE loss. At inference time, images can be generated by using the in-domain text encoder along the projection, with additional domain-related features passing through while keeping most of the CLIP latent space structure intact, so that the U-Net conditioning allows for clinically correct generated images.

Notice that the prompts the model is trained on can have an impact on the performance, that we try to measure: we explore object-oriented prompts of the form "a photo of a ..." and style-oriented prompts of the form "a photo in the style of a ...", with lexical variants of these two base prompts.

\subsection{Textual Embeddings Fine-tuning}
\label{textual-inversion}

Following the approach of \cite{https://doi.org/10.48550/arxiv.2208.01618}, the Stable Diffusion model can be fine-tuned to generate better looking images for the radiology setting by focusing on the embeddings of the text encoder. In this case, during training, the VAE, the U-Net, as well as all the other layers of the text encoder are frozen. In addition, a new token gets introduced, that can either describe: patient-level features, such as gender, age and body weight; procedure-level features, such as body part and modality; abnormality-level features, such as "no findings" or "pleural effusion".

As an example, we could introduce the token $<lung-xray>$ that is supposed to describe both a body part, lungs, and a modality, X-ray. This learning approach, denoted \textit{Textual Inversion}, zeroes out all the gradients associated with the embeddings of the already existing tokens, and in the end only learns the embedding of this newly introduced token.

Then, during training, generic input prompts with these new tokens are introduced, along associated radiology images. The rest is very similar to original training of the Stable Diffusion model, in that the model gets used to generate a synthetic image, and the noise at several timesteps in both the forward and backward process of the U-Net are passed through an MSE loss. Gradients are then used to only update the embeddings of the newly introduced tokens. 

\subsection{U-Net Fine-tuning}
\label{u-net-fine-tuning}

Finally, in a similar approach to \cite{https://doi.org/10.48550/arxiv.2208.12242}, one can improve the baseline Stable Diffusion model to generate better domain-specific images by relying on fine-tuning the U-Net. Instead of switching text encoders and using a projection (see Section \ref{textual-projection}) or training the embeddings of new tokens (see Section \ref{textual-inversion}), with this approach, all components except the U-net are kept frozen. In this sense, the setting is very similar to the approach of Section \ref{textual-projection}, except that no new token gets added, and the freezing is over the set of parameters of the U-Net. Then, the training is similar to the training of the original Stable Diffusion model, relying on MSE loss at several time steps of the denoising process to progressively converge to better generation of in-domain images.

\subsection{Classification of generated samples with a pretrained model}
After establishing the abovementioned methods, 50 samples per experimental setup were generated for the text prompts 'a photo of a lung xray' (negative sample) and 'a photo of a lung xray with visible pleural effusion' (positive sample), respectively. The generated images were then classified by the DenseNet-121 as detailed in Section \ref{variational-autoencoder}.

\section{Results}

\subsection{Training details}

Experiments were conducted on several devices predicated by computational needs of the methods. VAE and text encoder experiments were run locally, with Apple M1 Pro and M1 Max GPUs. Textual projection relied on 3 NVIDIA Quadro P5000 GPUs, with a single run taking 3 hours for 10k training steps in the case of document-level training, and 8 hours for 10k training steps in the case of token-level training, when using only one of these GPUs. Textual embedding fine-tuning and U-net fine-tuning used a NVIDIA V100 GPU and took respectively 1 hour for 3k training steps and 15 minutes for 400 training steps.

Model weights and implementations were obtained from the \textit{Hugging Face} platform \citep{https://doi.org/10.48550/arxiv.1910.03771}, notably making extensive use of the recently released \textit{diffusers} library \citep{von-platen-etal-2022-diffusers}. The Stable Diffusion weights were obtained from the \textit{CompVis/stable-diffusion-v1-4} repository. Weights of the in-domain text encoders were obtained using the sources referenced in the referenced publications.

\subsection{Variational autoencoder}

\begin{figure}[hbt!]
  \centering
    \includegraphics[width=135mm]{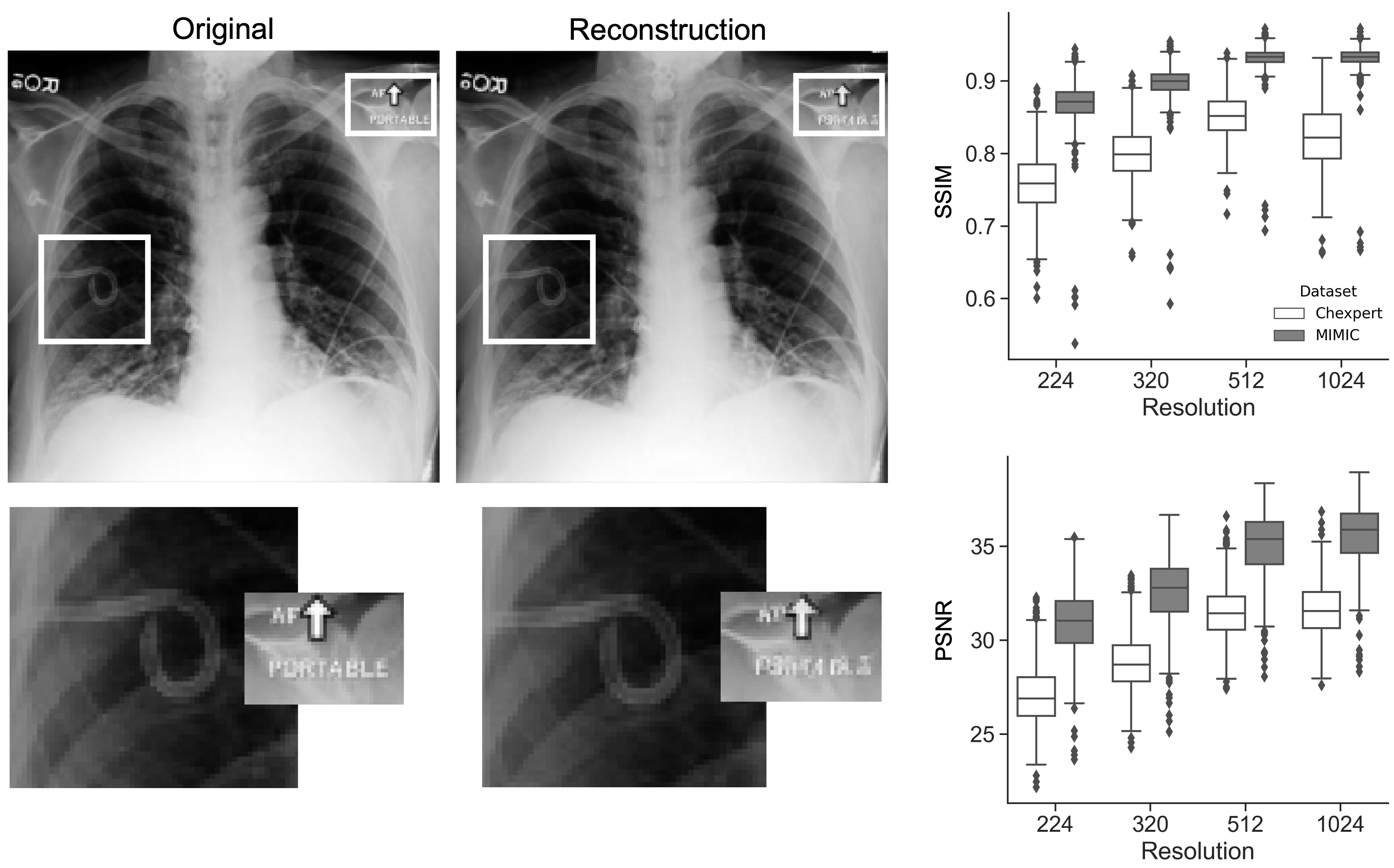}
  \caption{Image reconstruction analysis. (Right) Original and reconstructed image. The small burnt-in annotations in the top right corner get scrambled (seen in almost all samples), while the majority of other features (e.g., rib contours, devices) are well-preserved. (Left) Distribution of image quality metrics assessed for each image-reconstruction pair. SSIM: Structural similiarity index measure. PSNR: Peak signal-to-noise ratio.}
  \label{figure-image-quality}
\end{figure}

1000 CXR images from CheXpert and MIMIC were encoded and decoded using the pretrained VAE from the Stable-Diffusion-v1.4 pipeline. For samples from MIMIC-CXR, uantitative assessment showed a low reconstruction error (RMSE	31.8$\pm$6.5; median, 30.2; range, 21.4 - 70.1; PSNR, 35.1$\pm$1.6; median 35.3; range, 28.1 - 38.4 at 512x512 resolution) and a high structual similarity of original and reconstructed images (SSIM, 0.93$\pm$0.02; median, 0.93; range, 0.69 - 0.97). Image reconstruction quality was slightly lower for samples form CheXpert (RMSE, 47.8$\pm$7.4; median, 47.6; range, 26.2 - 75.5; PSNR, 31.5$\pm$1.4; median, 31.4; range, 27.4 -36.6; SSIM, 0.85$\pm$0.03; median, 0.85; range, 0.72 - 0.93), which might be attributed to different preprocessing procedures during the creation of the datasets, and did not affect visual analysis noticeably. See Figure \ref{figure-image-quality} for details. Image quality metrics did not depend on the class labels of the images (data not shown). 

Visual analysis was performed on 100 images sampled from MIMIC-CXR, resized to a maximum size of 768 pixels (height or width) and yielded a generally good perceived reconstruction quality (Mean visual score 4.51$\pm$0.54; median score, 5; range, 3 - 5). No reconstruction resulted in a completely non-diagnostic image (score 1) or altered the diagnostic information in a potentially problematic way (score 2). Almost all burnt-in text annotations were scrambled beyond recognition, however, diagnostic features were well preserved in almost all cases. Most of the score deductions were for blurred device components, cerclages and wires that couldn't be traced reliably after reconstruction, or blurred rib contours.

The reconstruction process slightly improved the classification performance of the DenseNet-121 model on CXR sampled from MIMIC-CXR for several categories, most notably for pleural effusion (AUC 0.881 vs. 0.877 for the original images), while the performance for other categories was negatively impacted (e.g., lung lesion, from AUC 0.766 in original images to 0.728 in reconstructed images). See Table \ref{mimic-recon-classification} for details. The same effect was 
only observed for the label 'Pleural Other' in the CheXpert sample, while most other labels were predicted with the same or slightly worse performance. 

The latent embeddings generated by the pretrained DenseNet-121 model were highly similar for image-reconstruction pairs for both MIMIC (mean cosine similarity, 0.98$\pm$0.02; median, 0.98; range, 0.88 - 1.00) and CheXpert (mean cosine similarity, 0.97$\pm$0.02; median, 0.98; range, 0.86 - 1.00).

\begin{table}[]
\caption{Classification performance of  DenseNet-121 for original and reconstructed CXR images from the MIMIC-CXR dataset. Enl Mediastinum = Enlarged Mediastinum.}
\label{mimic-recon-classification}
\centering
\begin{tabular}{@{}lccccccc@{}}
\toprule
\multirow{2}{*}{Finding} & \multirow{2}{*}{Prevalence} & \multicolumn{2}{c}{AUC} & \multicolumn{2}{c}{Accuracy} & \multicolumn{2}{c}{F1Score} \\
                  &                             & orig.   & recon.          & orig.        & recon.        & orig.        & recon.       \\ \midrule
Atelectasis       & 0.289                       & 0.805   & \textbf{0.807}  & 0.741        & 0.731         & 0.623        & 0.624        \\
Cardiomegaly      & 0.273                       & 0.810   & \textbf{0.812}  & 0.741        & 0.742         & 0.616        & 0.609        \\
Consolidation     & 0.119                       & 0.854   & \textbf{0.858}  & 0.815        & 0.818         & 0.496        & 0.494        \\
Edema             & 0.172                       & 0.858   & \textbf{0.862}  & 0.854        & 0.852         & 0.614        & 0.591        \\
Enl. Mediastinum  & 0.144                       & 0.790   & \textbf{0.793}  & 0.757        & 0.757         & 0.454        & 0.451        \\
Fracture          & 0.049                       & 0.651   & \textbf{0.653}  & 0.751        & 0.757         & 0.132        & 0.141        \\
Lung Lesion       & 0.077                       & 0.766   & 0.728           & 0.915        & 0.915         & 0.023        & 0.000        \\
Lung Opacity      & 0.297                       & 0.814   & 0.814           & 0.781        & 0.772         & 0.604        & 0.594        \\
Pleural Effusion  & 0.213                       & 0.877   & \textbf{0.881}  & 0.849        & 0.849         & 0.664        & 0.662        \\
Pleural Other     & 0.046                       & 0.780   & 0.778           & 0.677        & 0.665         & 0.161        & 0.177        \\
Pneumonia         & 0.226                       & 0.753   & 0.739           & 0.692        & 0.710         & 0.488        & 0.477        \\
Pneumothorax      & 0.030                       & 0.811   & 0.791           & 0.814        & 0.812         & 0.162        & 0.153        \\ \bottomrule
\end{tabular}
\end{table}

\subsection{Text Encoder}

Various text encoders and associated embedding methods are assessed on radiology reports in order to evaluate which method can retain optimum clinical knowledge in the latent representations. 

Following the definitions in section \ref{method-text-encoder} of the $text\_encoder$ models, the $extraction$ methods and the metric $CheXpert@k$, we compute in Table \ref{text-encoder-how-to-embed} for each model and each method the macro-average of the $CheXpert@k$ score aggregated per abnormality class, with $k=10$. As seen in the table, the method "CLS hidden state" is in general the one that works best to maximize the quality of the document-level representations of the impression sections. In addition, the model CXR-BERT-specialized is the one that reaches highest performance, taking for each model the corresponding $extraction$ method that worked best. 

Then, using the $extraction$ method that works best for each model, we can compute class-wise $CheXpert@k$ scores as well as the macro-averaged ones. These results are aggregated in Table \ref{text-encoder-cluster-impressions}. As a baseline, we use a bag-of-words approach that outputs a similarity measure between two reports using an intersection over union measure. This baseline does not create any embeddings, but provide a token-based similarity measure: we observe that the latent representations of the best models, on top of contracting the text space, better encode document-level content and result in higher scores. 

We remark that PubMedBERT, ClinicalBERT and CXR-BERT-general are three models that perform significantly less well than the other models, and should therefore, if possible, not be preferred for tasks that involve radiology reports. On the contrary, the two best performing models are CXR-BERT-specialized and the CLIP text encoder. As CLIP text encoder was not specifically trained on radiology reports, this underlines the quality of the training and the associated model. Using CXR-BERT-specialized instead would only improve performance by +15\%. 

For these reasons, we explore the textual projection with CXR-BERT-specialized, but also assess CLIP performance to be high enough to not justify replacing the text encoder in the various textual inversion and U-Net fine-tuning experiments. 

\begin{table}
  \caption{Macro-average of $CheXpert@10$ scores computed per abnormality class, over the impression sections of a set of radiology reports. Models that are better are retaining medical features get a higher score.}
  \label{text-encoder-how-to-embed}
  \centering
  \begin{tabular}{lllll}
    \toprule
    Model     
    & CLS hidden state
    & Mean hidden states
    & Pooler output
    & Model specific\\
    \midrule
    PubMedBERT
    & \textbf{30.8}
    & 23.6
    & 20.6
    & None\\
    ClinicalBERT
    & 26.3
    & \textbf{35.1}
    & 14.3
    & None\\
    SapBERT
    & \textbf{49.1}
    & 48.7
    & 41
    & None\\
    RadBERT
    & \textbf{54.2}
    & 32.8
    & 34.7
    & None \\
    CXRBERTgeneral
    & \textbf{32.4}
    & 25.4
    & 31.6
    & None\\
    CXRBERTspecialized
    & \textbf{61.1}
    & 34.5
    & None
    & 50.3\\
    ClipTextEncoder
    & 7.0
    & 42.8
    & \textbf{52.9}
    & None\\
    \bottomrule
  \end{tabular}
\end{table}

\begin{table}
  \caption{For each text encoder and the associated best $extraction$ method as computed in Table \ref{text-encoder-how-to-embed}, class-wise and macro-averaged $CheXpert@10$ scores are computed. Higher scores denote better capability at retaining important clinical features in the structure of the latent space.}
  \label{text-encoder-cluster-impressions}
  \centering
  \begin{tabular}{lllllllll}
    \toprule
    Abnormality     
    & Base
    & Pub.
    & Clin.
    & Sap.
    & Rad.
    & gen.
    & spe.
    & Clip.\\
    Atelectasis
    & 33.4
    & 21.8
    & 19.2
    & 54.2
    & 53
    & 23.4
    & \textbf{64.2}
    & 52.8\\
    Cardiomegaly
    & 21.6
    & 20.8
    & 10.2
    & 51
    & 53
    & 21.6
    & \textbf{67.2}
    & 47.6\\
    Consolidation
    & 36
    & 13.4
    & 35.8
    & \textbf{39.6}
    & 38
    & 35.4
    & 27
    & 38.4\\
    Edema
    & 62.8
    & 54
    & 62.6
    & 64.6
    & 67.2
    & 47.4
    & \textbf{85.4}
    & 72\\
    Enlarged Cardiomediastinum
    & 38
    & 21.2
    & 30.2
    & 41.8
    & \textbf{44.8}
    & 35.2
    & 37.6
    & 42.6\\
    Fracture
    & 49
    & 36.2
    & 35.6
    & 73.2
    & 72.6
    & 50.8
    & \textbf{83.2}
    & 74.2\\
    Lung Lesion
    & 30.2
    & 24
    & 21.2
    & 32
    & 37.8
    & 24.8
    & \textbf{56.2}
    & 33.8\\
    Lung Opacity
    & 20.4
    & 16.2
    & 20.6
    & 20.4
    & \textbf{34.2}
    & 20.4
    & 23.2
    & 25.6\\
    No Finding
    & 78.4
    & \textbf{82.2}
    & 75.4
    & 74.8
    & 79.8
    & 75.4
    & 76.8
    & 80.6\\
    Pleural Effusion
    & 46.4
    & 25
    & 39.4
    & 42.6
    & 65.8
    & 24.2
    & \textbf{72.2}
    & 68\\
    Pleural Other
    & 21.6
    & 13.6
    & 17.8
    & 36
    & 43.4
    & 16.8
    & \textbf{54}
    & 34.6\\
    Pneumonia
    & 53.8
    & 33.8
    & 40.4
    & 42.6
    & 44.4
    & 24
    & 45
    & \textbf{54}\\
    Pneumothorax
    & 56.4
    & 39.6
    & 60.6
    & 65.2
    & 73.6
    & 28.6
    & \textbf{92.8}
    & 72\\
    Support Devices
    & 32.6
    & 29.2
    & 23
    & 49.6
    & 50.8
    & 25.8
    & \textbf{70.4}
    & 44.8\\
    Macro
    & 41.5
    & 30.8
    & 35.1
    & 49.1
    & 54.2
    & 32.4
    & \textbf{61.1}
    & 52.9\\
    \bottomrule
  \end{tabular}
\end{table}

\subsection{Radiology Image Generation}

\begin{figure}
  \centering
    \includegraphics[width=140mm]{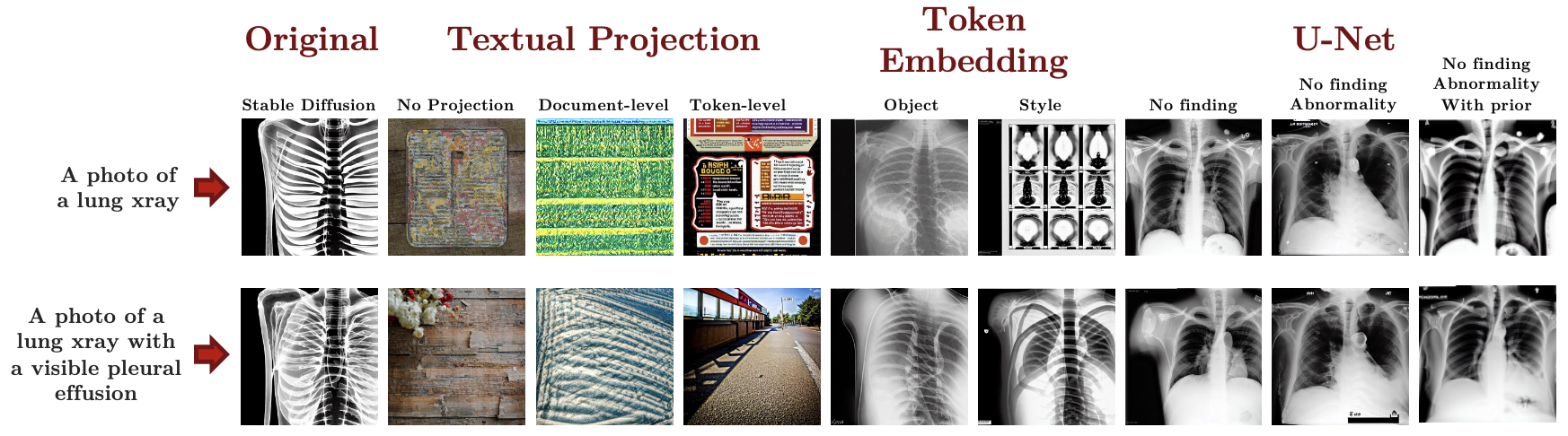}
  \caption{Images generated by various methods conditioned on radiology text prompts.}
  \label{figure-comparison-generation-methods}
\end{figure}

We tested the various generative models, after potential fine-tuning by one of the scenarios described in Section \ref{materials-methods}, with two simple prompts: "A photo of a lung xray" and "A photo of a lung xray with a visible pleural effusion". Based only on this text-conditioning, the models generated synthetic images. We used the Fréchet inception distance as introduced in Section \ref{variational-autoencoder} to measure the quality of the generated images. The results are compiled in Table \ref{image-generation-table}, along an empirical sample of images as produced by each method in Figure \ref{figure-comparison-generation-methods}.

For the most simple prompt "A photo of a lung xray", the baseline model, the Stable Diffusion model as released by \cite{Rombach2022}, generates black-and-white thoracic images, visible on Figure \ref{figure-comparison-generation-methods}, with associated FID-score of 0.097. Fine-tuning the stable diffusion model using the textual projection approach leads to a degradation of the generated images, with FID-scores that at least double for both the token-level and the document-level projections. Textual- inversion-based fine-tuning do not degrade performance much but do not allow for an improvement of the FID-score as well. Only with the method that consists in fine-tuning the U-net do we observe enhanced generated images, lowering the FID-score down to 0.034. For more complex prompts such as "A photo of a lung xray with a visible pleural effusion", the stable diffusion baseline shows even more limitations, with larger FID-score 0.151, being outperformed by both textual inversion and U-Net fine-tuning.

The textual projection does not seem to converge well enough: samples from Figure \ref{figure-comparison-generation-methods} shows the generated images to be out-of-domain. Nevertheless, we estimate that a more complex architecture, instead of our simple 1-hidden-layer projection, could be worth exploring: if projection-based domain-adaptation turns out to produce interesting examples, this could open the door to quick domain-adaptation for the large amount of pre-trained text encoders that are now available.

Out of all the methods, the U-Net fine-tuning seems by far the most promising: it gets the lowest FID-scores and obviously the most realistic outputs. Nevertheless, we also notice in this setting the limitations of our non-medical-based metric: samples clearly show that U-Net fine-tuning with prior enables the model to learn the difference between "no findings" and "pleural effusion", something a model trained without a prior can not do. As seen in Table \ref{image-generation-table}, FID fails to capture this improvement. We assess that further progress in the domain-specific generation of images for radiology require the use of more domain-specific metrics, that would be able to capture the ability of the model to correctly insert abnormalities that are coherent with the conditioning text prompt.

\vspace{2em}

\begin{table}[bt!]
\caption{Evaluation of the quality of generated images with different methods for adapting stable diffusion to the radiology domain. Scores represent the Fréchet inception distance (FID), and lower scores mean better generated images.}
\label{image-generation-table}
\centering
\resizebox{\linewidth}{!}{%
\begin{tabular}{lccc} 
\toprule
Training Strategy
& \begin{tabular}[c]{@{}c@{}}A photo of\\a lung xray\end{tabular}
& \begin{tabular}[c]{@{}c@{}}A photo of a lung xray\\with a visible pleural effusion\end{tabular}
& \begin{tabular}[c]{@{}c@{}}A photo in the style of\\a lung xray\end{tabular}
\\ 
\midrule
\textit{Original model}
&
&
&
\\
~ ~ Stable Diffusion
& 0.097
& 0.151
&
\\
\midrule
\textit{Textual Projection}
&
&
&
\\
\textit{CXR-BERT-specialized}
&
&
&
\\
~ ~ No Projection
& 0.124
& 0.144
& 
\\
~ ~ Document-level projection
& 0.266
& 0.104
& 
\\
~ ~ Token-level projection
& 0.201
& 0.257
& 
\\
\midrule
\textit{Token embedding training}
&
&
&
\\
~ ~ Object, radiology
& 0.108
& 0.058
& 0.092
\\
~ ~ Object, lung
& 0.135
& 
& 0.135
\\
~ ~ Style, radiology
& 0.101
& 0.057
& 0.084
\\
~ ~ Style, lung
& 0.130
& 
& 0.083
\\
\midrule
\textit{U-Net training}
&
&
&
\\
~ ~ Trained on no findings
& 0.057
& 0.043
& 
\\
~ ~ Trained on no findings and abnormality
& \textbf{0.034}
& \textbf{0.041}
&
\\
~ ~ Trained on no findings and abnormality with prior
& 0.170
& 0.086
& 
\\
\bottomrule
\end{tabular}
}
\end{table}

\begin{figure}
  \centering
    \includegraphics[width=135mm]{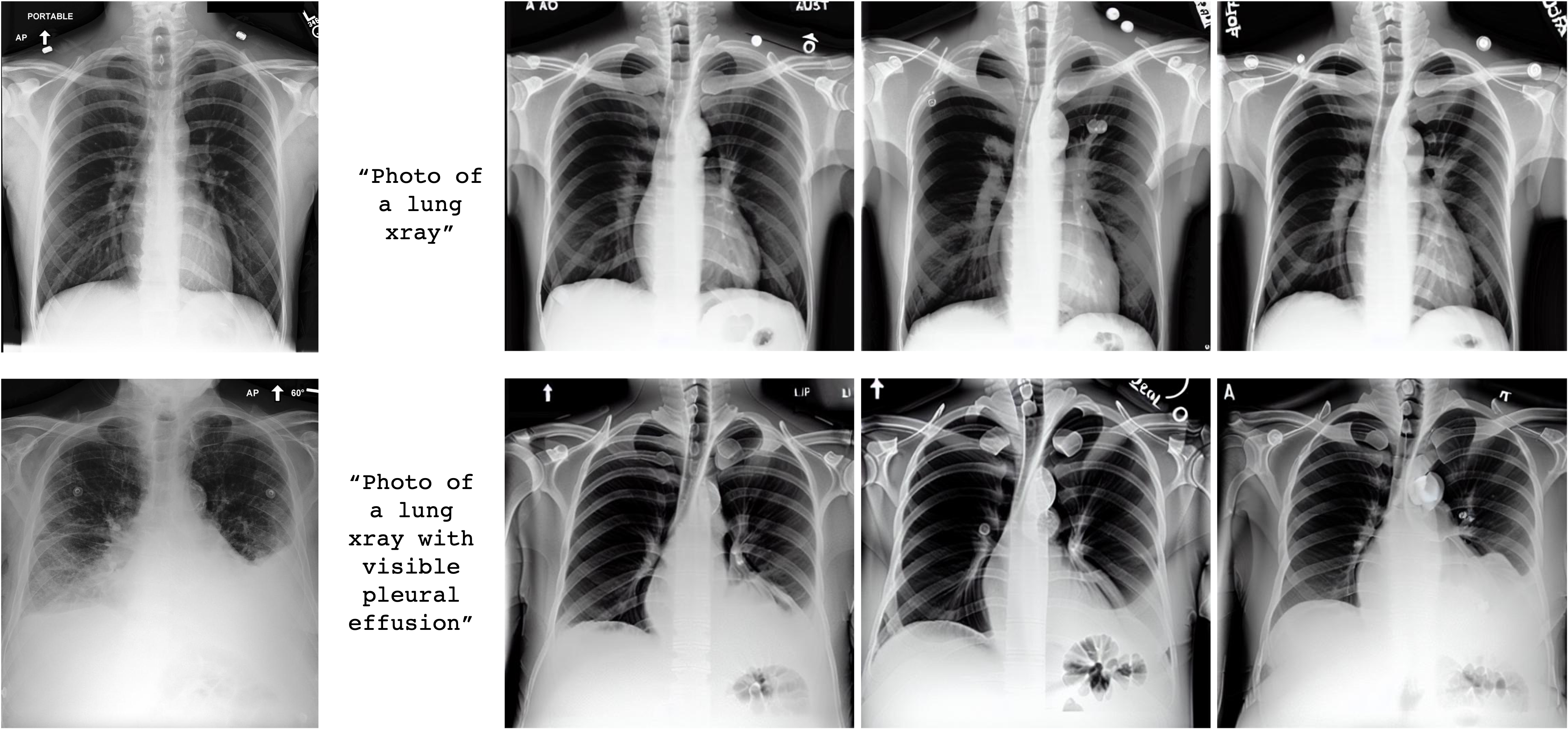}
  \caption{Original CXRs used for fine-tuning the model (on the left) and synthetic CXRs generated by text-conditioning of the U-net fine-tuned Stable Diffusion pipeline (on the right).}
  \label{figure-realvsgeneratedgallery}
\end{figure}


\subsection{Classification of synthetic CXR}

100 samples of synthetically generated CXR (with 50 samples text-conditioned to show a pleural effusion) were classified by the DenseNet-121 model for the presence or absence of pleural effusion. As heralded by visual analysis, a baseline in the form of the original Stable Diffusion pipeline failed to produce meaningful signal for a CXR classification model with an AUC of 0.5. Textual projection significantly decreased the classification performance for document-level representations as detailed in Section \ref{textual-projection} and only slightly increased the performance for token-level representations. Textual inversion, as explained in Section \ref{textual-inversion} increased the classification performance to a moderate AUC of 0.74 with a high precision and low recall. Fine-tuning the U-Net component without prior as detailed in Section \ref{u-net-fine-tuning} slightly decreased the classification performance compared to the baseline, which is attributed to 'forgetting' during the fine-tuning process. Fine-tuning the U-Net with prior significantly outperformed the baseline with a nearly perfect AUC of 0.98. For details see Table \ref{table-generated-classification}.

\begin{table}[]
\centering
\caption{Classification performance of DenseNet-121 on 100 samples with and without text conditioning towards displaying a pleural effusion. The text prompt was 'a photo of a lung xray' for negative samples, and 'a photo of a lung xray with visible pleural effusion' for positive samples.}
\label{table-generated-classification}
\begin{tabular}{@{}lllllll@{}}
\toprule
Method                      & Prevalence & AUC & Accuracy & F1Score & Precision & Recall \\ \midrule
Original Model               & 0.5        & 0.514 & 0.500    & 0.167   & 0.500     & 0.100  \\
Textual Projection (Doc.) & 0.5        & 0.136 & 0.480    & 0.000   & 0.000     & 0.000  \\
Textual Projection (Token) & 0.5        & 0.582 & 0.540    & 0.489   & 0.550     & 0.440  \\
Textual Inversion            & 0.5        & 0.742 & 0.610    & 0.381   & 0.923     & 0.240  \\
U-Net, no prior         & 0.5        & 0.459 & 0.500    & 0.000   & 0.000     & 0.000  \\
U-Net, with prior       & 0.5        & \textbf{0.984} & 0.950    & 0.947   & 1.000     & 0.900  \\ \bottomrule
\end{tabular}
\end{table}

\section{Discussion}

Following our experiments, we were able to assess the recently released Stable Diffusion model, including its VAE, U-Net and CLIP text encoder components, and their capacity to produce medically realistic images based on text prompts that describe observable abnormalities. We conducted quantitative and qualitative evaluations, showing that: the VAE is powerful enough to preserve and reconstruct radiological CXR images without specific training, including  abnormalities and clinically relevant features; the CLIP text encoder accurately represents simple radiology-specific text prompts, outperforming 4 out of the 6 reviewed domain-specific text encoders. We proposed and explored textual projection, a domain-adaptation method, textual inversion and U-Net fine-tuning, and, with the latter, obtained a model capable of generating synthetic CXR images that are visually and quantitatively outperforming the baseline, and that can correctly represent abnormalities. 

No study has previously explored latent diffusion models for the generation of synthetic CXRs or other modalities relevant to thoracic imaging. Additionally, this is the first study to systematically explore the capabilties of the recently published Stable Diffusion pipeline for the generation of medical images.

Latent diffusion models that form the foundation of the Stable Diffusion pipeline have recently been explored for the generation of synthetic brain MRI images, with the ability to condition the generative process on variables like age, sex and brain structure volumes \citep{pinaya2022ldmbrain}. In contrast to this, our work presents a multi-modal approach that uses natural language text conditioning, which covers and is able to convey a much broader spectrum of information used for medical decision making than class labels. This multi-modal approach also allows us to analyze and verify the obtained results on multiple levels: human text comprehension for the designed prompts and their alignment with the findings in the generated images; quantitative image analysis through metrics; perceived image quality analysis through inspection; and evaluation of deep neural network  performance by analysis of the latent representations as well as the classification results for the intended pathologies. Notably, this is achieved in a comparably low-resource setting, by leveraging powerful pretrained multi-modal models forming the Stable Diffusion pipeline. As explored in and verified by this work, the generative process is exceedingly flexible, as the conditioning can be performed on arbitrary inputs, in contrast to being confined to certain environments, scanner types, or modalities.

The findings detailed in this work can serve as a foundation to further explore the potential of latent diffusion-based models to learn a wide-range of abnormalities, the ability to combine them, as well as extending the research to other imaging modalities and body parts. 

A limitation of our approach is that the employed metrics have limited capacity to assess the clinical correctness of the generated images, which we adressed by visual evaluation by a trained radiologist. In addition, our fine-tuned stable diffusion model lacks diversity in the images it generates, an important attribute of generative models. We attribute this to the small number of samples that were used to fine-tune the U-Net component in a few-shot manner. Finally, the text prompts the models are conditioned on are synthetic and do not fully correspond to the wording used in the clinical setting, so that models capable of being conditioned on entire or partial radiology reports are an area of future research.

\section{Conclusion}

In this paper, we studied the recently released Stable Diffusion model and its performance when being conditioned on domain-specific text prompts to generate synthetic radiological images. First, we determined that the variational autoencoder of the Stable Diffusion model is capable of reconstructing chest X-ray images while sufficiently preserving the medical features of interest, obviating the need for radiology-specific fine-tuning. Second, we compared the CLIP text-encoder used for Stable Diffusion conditioning to several in-domain text-encoders, showing that CLIP generates text-embeddings that preserve abnormality-related content, being outperformed by no more than 15\% by the best in-domain text-encoder on the $CheXpert@10$ metric. Third, we developed and explored a fine-tuning strategy, textual projection, that replaces the CLIP text-encoder with an in-domain text-encoder and train a projection from the in-domain latent space to CLIP text latent space, trying to leverage added knowledge from the in-domain text encoder. It led to a degradation of quality and out-of-domain samples, compared to the baseline Stable Diffusion model. Fourth, we explored the textual inversion approach and showed it can learn radiology-specific visual features, though struggling at distinguishing between abnormalities. Fifth, we generated high-fidelity CXR images by fine-tuning the U-net component of the Stable Diffusion model, demonstrating that such generative models can learn radiology concepts and be used to insert realistic-looking abnormalities. Future research will focus on addressing the lack of diversity of the generated images, on expanding the representational capacity to a wider range of medical concepts and modalities, as well as on enhancing the text conditioning with more realistic clinical prompts.  

\section{Acknowledgements}

Research reported in this publication was made possible in part by the \textit{National Institute of Biomedical Imaging and Bioengineering (NIBIB)} of the \textit{National Institutes of Health}, which funded PC under contracts
75N92020C00008 and 75N92020C00021. CB received support from the Swiss Society of Radiology and the Kurt and Senta Herrmann-Foundation, unrelated to this work. We acknowledge the support of this work by the Wu Tsai Human Performance Alliance at Stanford University and the Joe and Clara Tsai Foundation.

{
\small
\bibliography{neurips_2022}
}

\appendix

\newpage
\section{Image reconstruction on CheXpert dataset}
\label{sec:appendix}

\begin{table}[ht!]
\caption{Classification performance of  DenseNet-121 for original and reconstructed CXR images from the CheXpert dataset. Enl Mediastinum = Enlarged Mediastinum.}
\label{chexpert-recon-classification}
\centering
\begin{tabular}{@{}lccccccc@{}}
\toprule
\multirow{2}{*}{Finding}          & \multirow{2}{*}{Prevalence} & \multicolumn{2}{c}{AUC} & \multicolumn{2}{c}{Accuracy} & \multicolumn{2}{c}{F1Score} \\
                           &                             & orig.   & recon.          & orig.        & recon.        & orig.        & recon.       \\ \midrule
Atelectasis                & 0.455                       & 0.693   & 0.690           & 0.532        & 0.550         & 0.653        & 0.661        \\
Cardiomegaly               & 0.211                       & 0.760   & 0.756           & 0.410        & 0.422         & 0.406        & 0.407        \\
Consolidation              & 0.277                       & 0.731   & 0.730           & 0.456        & 0.477         & 0.493        & 0.499        \\
Edema                      & 0.335                       & 0.766   & 0.766           & 0.467        & 0.486         & 0.549        & 0.555        \\
Enl. Mediastinum & 0.187                       & 0.601   & 0.588           & 0.289        & 0.305         & 0.325        & 0.326        \\
Fracture                   & 0.087                       & 0.568   & 0.536           & 0.102        & 0.119         & 0.158        & 0.159        \\
Lung Lesion                & 0.086                       & 0.614   & 0.587           & 0.232        & 0.247         & 0.172        & 0.168        \\
Lung Opacity               & 0.554                       & 0.730   & 0.723           & 0.648        & 0.661         & 0.748        & 0.751        \\
Pleural Effusion           & 0.420                       & 0.818   & 0.813           & 0.597        & 0.607         & 0.663        & 0.664        \\
Pleural Other              & 0.057                       & 0.550   & \textbf{0.556}  & 0.724        & 0.687         & 0.110        & 0.103        \\
Pneumonia                  & 0.203                       & 0.668   & 0.663           & 0.371        & 0.370         & 0.375        & 0.374        \\
Pneumothorax               & 0.092                       & 0.674   & 0.658           & 0.180        & 0.209         & 0.178        & 0.179        \\ \bottomrule
\end{tabular}
\end{table}

\section{Text-encoder clusterization capabilities for various chest abnormalities}

Following the text-encoder comparison, to determine which models best retain radiology features, we fed the impression sections of the set of MIMIC radiology reports introduced in Section \ref{section-datasets} to each candidate text-encoder. Then, using a t-SNE projection \citep{JMLR:v9:vandermaaten08a}, we visualized the clusters per abnormality in the embedding space of each text-encoder, as seen on the following figures.

\begin{figure}[hbt!]
  \centering
    \includegraphics[width=130mm]{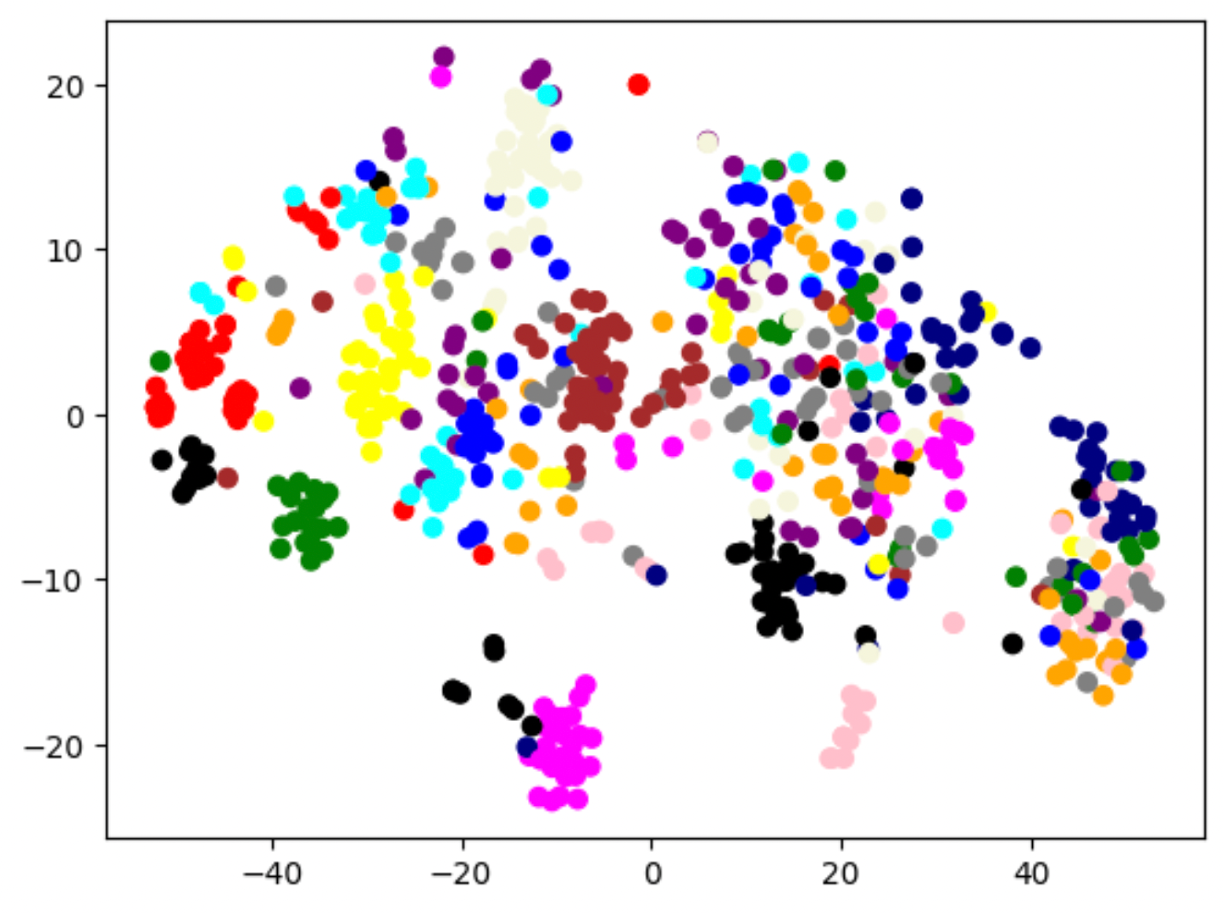}
  \caption{t-SNE projection of the CLIP embeddings for various radiology reports with different abnormalities. Each abnormality corresponds to a color.}
  \label{clusters-clip}
\end{figure}

\begin{figure}[hbt!]
  \centering
    \includegraphics[width=130mm]{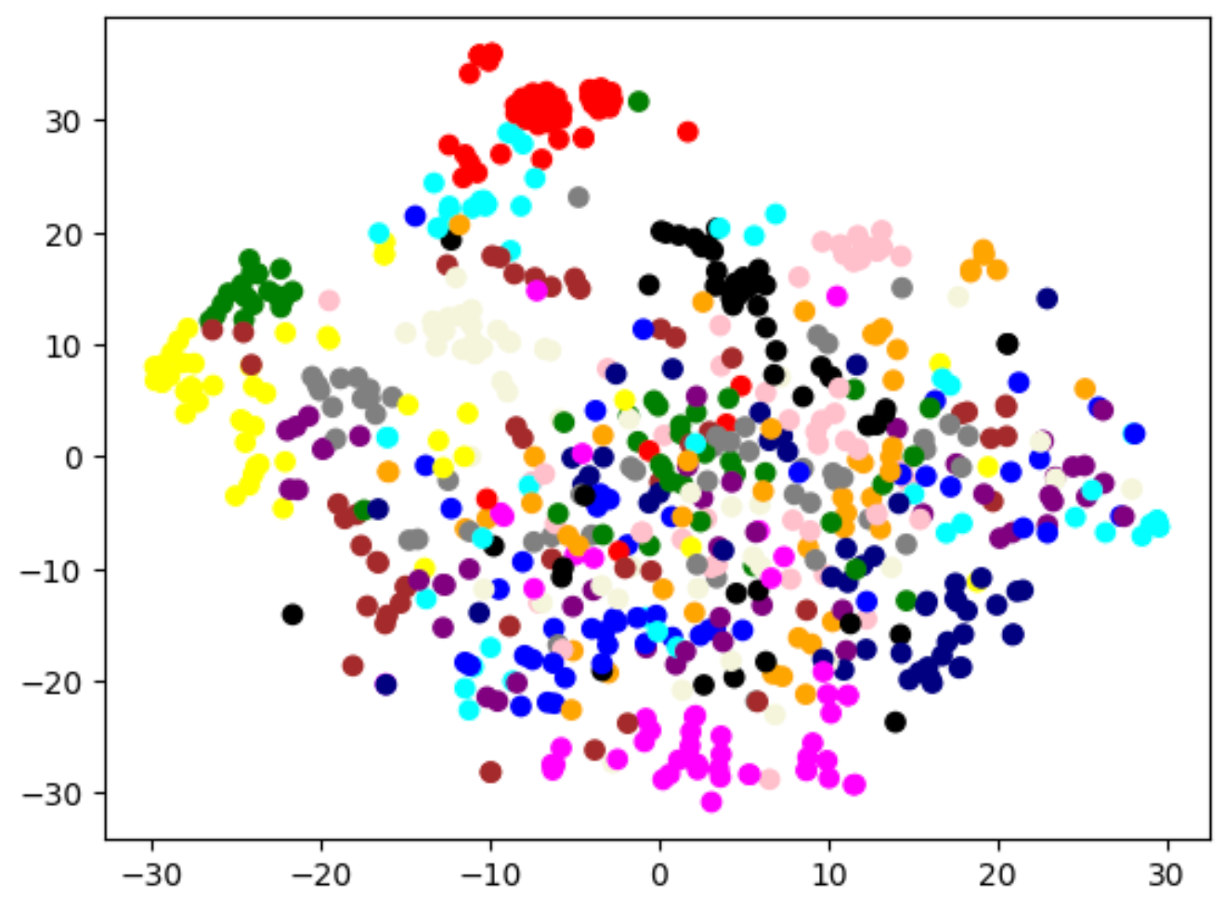}
  \caption{t-SNE projection of the SapBERT-from-PubMedBERT-full embeddings for various radiology reports with different abnormalities. Each abnormality corresponds to a color.}
  \label{clusters-sap}
\end{figure}

\begin{figure}[hbt!]
  \centering
    \includegraphics[width=130mm]{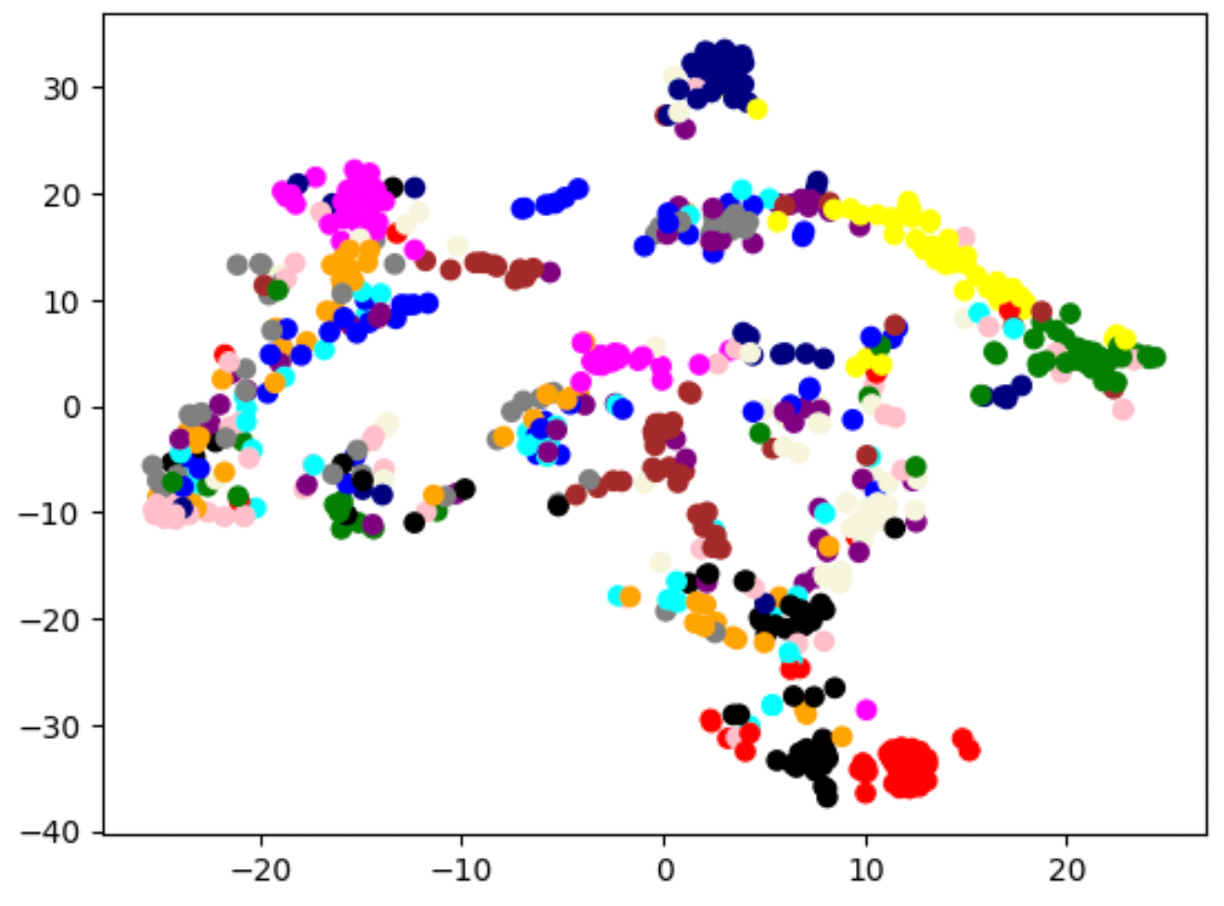}
  \caption{t-SNE projection of the CXR-BERT-specialized embeddings for various radiology reports with different abnormalities. Each abnormality corresponds to a color.}
  \label{clusters-microsoft}
\end{figure}

For the reference, each color in the preceding figures describe the following abnormalities: beige Atelectasis; green Cardiomegaly; blue Consolidation; yellow Edema; pink Enlarged Cardiomediastinum; black Fracture; orange Lung Lesion; purple Lung Opacity; red No Finding; brown Pleural Effusion; gray Pleural Other; cyan Pneumonia; magenta Pneumothorax; navy Support Devices.

\end{document}